\pdfoutput=1
\documentclass[11pt]{article}

\usepackage[preprint]{acl}
\usepackage{setspace}
\usepackage{times}
\usepackage{latexsym}
\usepackage{siunitx}
\usepackage[T1]{fontenc}
\usepackage[utf8]{inputenc}

\usepackage{microtype}
\usepackage{amssymb}
\usepackage{inconsolata}

\usepackage{graphicx}
\usepackage{enumitem}
\usepackage{hyperref}
\usepackage{url}
\usepackage{booktabs} 
\usepackage{colortbl}
\usepackage[most]{tcolorbox}
\usepackage{adjustbox}
\usepackage{multirow}
\PassOptionsToPackage{table}{xcolor}
\usepackage{arydshln}
\usepackage[table]{xcolor}
\usepackage{pgfplots}
\definecolor{GPT4oColor}{RGB}{65, 156, 167}
\definecolor{GPT4oMiniColor}{RGB}{67, 156, 67}
\definecolor{Qwen2-7BColor}{RGB}{227, 150, 149} 
\definecolor{Qwen2-72BColor}{RGB}{67, 204, 204} 
\definecolor{Llama3.1-8BColor}{RGB}{204, 67, 204} 
\definecolor{Llama3.1-70BColor}{RGB}{210, 210, 99} 
\definecolor{Llama3.1-405BColor}{RGB}{65, 65, 65} 
\definecolor{ConstraintColor}{RGB}{252, 141, 98} 
\definecolor{QuestionColor}{RGB}{102, 194, 165}   
\definecolor{EpisodicColor}{RGB}{141, 160, 203}  
\definecolor{caribbeangreen}{rgb}{0.0, 0.8, 0.6}

\title{Revealing the Barriers of  Language Agents in Planning}

\author{
    Jian Xie\textsuperscript{\rm $\spadesuit$} \quad
    Kexun Zhang\textsuperscript{\rm $\heartsuit$}\thanks{Equal Contribution} \quad
    Jiangjie Chen\textsuperscript{\rm $\diamondsuit$}\footnotemark[1]\thanks{Part of the work done while at Fudan University.}  \quad  
    Siyu Yuan\textsuperscript{\rm $\spadesuit$}  \quad \\
    \bf Kai Zhang\textsuperscript{\rm $\clubsuit$} \quad
    Yikai Zhang\textsuperscript{\rm $\spadesuit$}  \quad
    Lei Li\textsuperscript{\rm $\heartsuit$}  \quad
    Yanghua Xiao\textsuperscript{\rm $\spadesuit$}
    \\
\textsuperscript{\rm $\spadesuit$}Fudan University \quad
\textsuperscript{\rm $\heartsuit$}Carnegie Mellon University \\
\textsuperscript{\rm $\diamondsuit$}ByteDance Inc. \quad
\textsuperscript{\rm $\clubsuit$}The Ohio State University\\
\small{\texttt{\{jianxie22, syyuan21, ykzhang22\}@m.fudan.edu.cn, kexun@cmu.edu}}
\\
\small{\texttt{jiangjiec@bytedance.com, zhang.13253@osu.edu, leili@cs.cmu.edu, shawyh@fudan.edu.cn
}
}}

\begin{document}
\maketitle
\begin{abstract}
Autonomous planning has been an ongoing pursuit since the inception of artificial intelligence. 
Based on curated problem solvers, early planning agents could deliver precise solutions for specific tasks but lacked generalization. 
The emergence of large language models (LLMs) and their powerful reasoning capabilities has reignited interest in autonomous planning by automatically generating reasonable solutions for given tasks.
However, prior research and our experiments show that current language agents still lack human-level planning abilities.
Even the state-of-the-art reasoning model, OpenAI o1, achieves only 15.6\% on one of the complex real-world planning benchmarks.
This highlights a critical question: \textit{\textbf{What hinders language agents from achieving human-level planning?}} 
Although existing studies have highlighted weak performance in agent planning, the deeper underlying issues and the mechanisms and limitations of the strategies proposed to address them remain insufficiently understood.
In this work, we apply the feature attribution study and identify two key factors that hinder agent planning: the \textbf{limited role of constraints} and the \textbf{diminishing influence of questions}.
We also find that although current strategies help mitigate these challenges, they do not fully resolve them, indicating that agents still have a long way to go before reaching human-level intelligence.
Resources are available on the \href{https://github.com/hsaest/Agent-Planning-Analysis}{GitHub}.
\end{abstract}

\section{Introduction}
\label{sec:intro}
Planning is the process of determining the sequence of actions needed to achieve a goal. 
It involves goal decomposition, constraint consideration, and foresight for simulating and predicting outcomes.
In the development of artificial intelligence, this capability is considered the ``Holy Grail'' for achieving or even surpassing human intelligence~\citep{kahneman2011thinking,openai2023planning}. 
However, the path to achieving autonomous planning is a long journey. 
Researchers have long focused on building custom systems tailored to specific tasks~\citep{newell1959report,mcdermott1992robot,silver2017mastering}. 
While these systems could deliver precise solutions through rigorous problem solvers, the extensive effort required for task-specific design prevents them from achieving universal problem-solving capabilities or general intelligence.

\begin{figure}[t]
\centering
    \includegraphics[width=\linewidth]{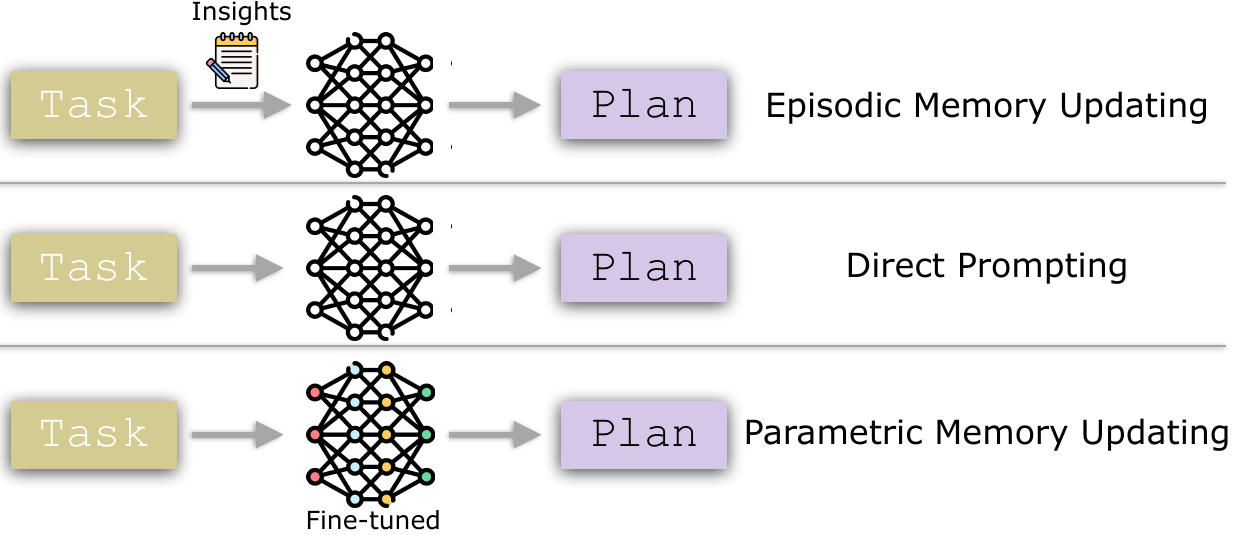}
    \caption{Memory updating strategies for language agents. Insights are learned from previous attempts.}
    \label{fig:fig1}
\end{figure}

The advent of language agents~\citep{weng2023prompt,su2023language,sumers2024cognitive}, which are powered by large language models (LLMs;~\citet{openai2022chatgpt,openai2023gpt4,team2023gemini,dubey2024llama,yang2024qwen2}), changes the landscape. 
Thanks to the flexibility of natural language, LLM-based language agents have shown strong potential to generalize to various planning tasks without relying on traditional curated, task-specific solvers written in domain-specific languages like Planning Domain Definition Language (PDDL).
However, despite these language agents demonstrating impressive capabilities across various tasks~\citep{yao2022react,yao2023tree,zheng2023seeact,gu2024middleware}, their performance in planning remains disappointing and is viewed as mere ``\textit{approximate retrieval}''~\citep{kambhampatiposition} rather than engaging in genuine reasoning.
Specifically, even the most capable model, OpenAI o1~\citep{openai2024o1}, which claims to surpass human PhD-level accuracy on several reasoning tasks, achieves only 15.6\% in a real-world travel planning benchmark, TravelPlanner (see Figure \ref{fig:main_res}), far below human-level planning abilities. 
To uncover the fundamental reasons behind the weak performance, we seek to answer the first research question in this paper: \textbf{\textit{RQ1: Why do current language agents struggle with planning?}}

In order to enhance language agents’ performance in planning tasks, numerous strategies have been proposed recently, which can be categorized into three main branches, as shown in Figure \ref{fig:fig1}: episodic memory updating through prompt optimization~\citep{zhao2024expel,shinn2024reflexion,fu2024autoguide}, parametric memory updating through model training~\citep{zeng2023agenttuning,song-etal-2024-trial,yin-etal-2024-agent}, and translating queries into formal planning languages, followed by resolution using external solvers~\citep{liu2023llmp,dagan2023dynamic}. 
Although these strategies have shown performance improvements across various tasks, their underlying mechanisms remain largely opaque. 
Moreover, these strategies still fall short of human-level intelligence~\citep{valmeekam2024planbench,valmeekam2024llms,stechly2024self}, particularly in complex real-world tasks~\citep{xietravelplanner,gundawar2024robust,chen2024can}.
Therefore, based on the findings from RQ1, this paper seeks to answer the research questions, \textbf{\textit{RQ2: What happens during memory updating for language agents}} and \textbf{\textit{RQ3: What hinders these strategies from achieving high-level planning abilities?}}
Specifically, we focus on language agents' vanilla planning as well as planning following memory updating, which reflect the internal planning capabilities of language agents rather than the translation ability.

In this paper, we delve into the two main components of planning: constraints and questions, which serve as the foundational elements for planning tasks. 
Constraints refer to the rules that agents must adhere to when generating a plan, while questions represent the goals that drive the planning process. 
Understanding how agents handle these elements is crucial for improving their performance in complex planning tasks.
Using Permutation Feature Importance~\citep{breiman2001random,fisher2019all} to analyze the feature attribution of constraints and questions, our investigation reveals several key findings:
\textit{\textbf{1)}} Language agents show a limited understanding of constraints, and the influence of the question weakens as the planning horizon increases.
\textit{\textbf{2)}} Episodic memory updating improves constraint understanding but relies on global understanding, and it's still difficult for agents to reference constraints in a fine-grained manner.
\textit{\textbf{3)}} Parametric memory updating enhances the question's impact on the final plan, but the diminishing influence of the question remains a challenge.
\textit{\textbf{4)}} Both strategies resemble ``shortcut learning'' and struggle with dynamic constraints in planning.

\section{Related Work}
\label{sec:related}
\subsection{Language Agent}
The advent of large language models sparks widespread attention due to their remarkable abilities, such as mathematical reasoning, creative writing, and information retrieval~\citep{gomez2023confederacy,Zhang2023LLM-QA4RE,Lou2023MUFFIN,zhu2024deductive}. 
Building on these models, language agents expand their capabilities to engage with the real world, including utilizing tools~\citep{gu2024middleware}, grounding environments~\citep{zheng2023seeact}, and even controlling real-world robotics~\citep{zeng2023large}, functioning as a ``reasoning brain'' beyond mere text generation.
The conceptual framework of language agents includes:
\textit{1)} \textbf{Memory module} handles both long-term memory embedded in the model's parameters, such as commonsense~\citep{west-etal-2022-symbolic}, and short-term memory specific to tasks~\citep{majumder2023clin}.
\textit{2)} \textbf{Tool-use module} enables agents to utilize external tools to compensate for inherent limitations, such as calling a calculator for arithmetic tasks or retrieving up-to-date information from external databases~\citep{lu2023chameleon,xie2024adaptive,wu2024how}.
\textit{3)} \textbf{Planning module} controls the entire task process, including goal decomposition, action sequencing, and forward estimation, requiring comprehensive and advanced reasoning abilities~\citep{weng2023prompt,sumers2024cognitive}.

\subsection{Planning in Language Agents}
Planning, a hallmark of human intelligence, serves as a critical component in language agent systems, as it directly controls task execution and goal achievement. 
Improving an agent's planning abilities thus leads to overall improvements across various tasks.
However, previous studies show that current agents still struggle with planning tasks, such as classical tasks like block manipulation~\citep{valmeekam2024planbench} or real-world tasks like travel planning~\citep{xietravelplanner,zhang-etal-2024-timearena}.
While these studies highlight agents' weaker performance in planning, they mainly provide high-level observations, leaving the deeper, underlying reasons less explored.
Furthermore, although strategies such as updating episodic memory (also referred to as working memory~\citep{zhao2024expel}), which allows learning from past trials and errors~\citep{shinn2024reflexion,fu2024autoguide}, or improving parametric memory through fine-tuning~\citep{yin-etal-2024-agent} have been proposed, the mechanisms driving these performance improvements, as well as their limitations, remain unclear.
Therefore, in this work, we aim to understand the challenges faced by current language agents in planning and provide promising directions for addressing weaknesses in planning strategies to guide the development of more effective agents.

\subsection{Interpretability of Language Models}
Despite the impressive capabilities of LLMs, their thinking processes remain opaque. 
Interpreting these models is essential for improving their reliability and transparency in real-world applications. 
Attention visualization helps explore how models allocate attention across different input elements, attention heads, and layers within the model~\citep{katz2023visit,zheng2024attention,luo2024understanding}.
Additionally, feature attribution methods analyze the importance of each input feature using techniques like perturbation~\citep{ribeiro2016should,fisher2019all} and gradients~\citep{sundararajan2017axiomatic,mudrakarta2018did}.
However, much of the existing work focuses on traditional tasks such as classification, which do not fully reflect the complexity of planning tasks. 
Planning requires handling long-horizon dependencies and balancing multiple objectives or constraints, presenting unique challenges that remain underexplored. 
To address this gap, this work utilizes interpretability techniques to investigate why agents struggle with planning tasks.

\section{Background}
\label{sec:related}
\subsection{Dataset}
We choose \textbf{Blocksworld} and \textbf{TravelPlanner} as our testbeds, which cover both classical planning and real-world complex planning scenarios:

\begin{itemize}

\item  \textbf{BlocksWorld}~\citep{valmeekam2024planbench} is a planning benchmark that provides a domain description, including action and constraint definitions, and requires agents to execute actions to transition from an initial state to a goal state. All actions must adhere to the explicit constraints outlined in the prompt.

\item  \textbf{TravelPlanner}~\citep{xietravelplanner} is a real-world travel planning benchmark that requires language agents to generate plans based on provided information and user queries, aligning with commonsense and the hard constraints specified in the queries. Unlike the static nature of BlocksWorld, the hard constraints in TravelPlanner are \textit{dynamic}, as they need to be inferred from the query and satisfied through item selection. We use the ``sole-planning'' mode to focus on the agents' planning ability, excluding the influence of tool-use abilities required in the ``two-stage'' mode.

\end{itemize}

\subsection{Permutation Feature Importance}
Permutation Feature Importance~\citep{breiman2001random,fisher2019all} is a strategy for evaluating the importance of features in a system. 
Specifically, if a feature is important, its removal or alteration will significantly affect the system’s result, whereas an unimportant feature will have little to no impact. 
In this paper, we adopt Permutation Feature Importance as our analysis strategy for testing the inner workings of language agents when planning.

Formally, given a language model \( P_{\theta} \), a feature sequence \( X = \{x_{1}, x_{2}, \ldots, x_{n}\} \), and a target sequence \( Y = \{y_{1}, y_{2}, \ldots, y_{m}\} \), the \textbf{attribution score} \( S_{i,j} \) for the contribution of feature \( x_{i} \) to the target \( y_{j} \) is defined as follows\footnote{For simplicity, we omit transformations like log and softmax here.}:
\begin{equation}
\small
S_{i,j} = {P_{\theta}(y_{j} \mid X, Y_{1:j-1})} - {P_{\theta}(y_{j} \mid \Hat{X}_i, Y_{1:j-1})}.
\end{equation}
A low or near-zero $S_{i,j}$ indicates that the feature \(x_{i}\) is almost independent of the target, while a higher score suggests a stronger contribution. 
Here, \( P_{\theta}(y_{j} \mid X, Y_{1:j-1}) \) represents the conditional probability of the target \( y_{j} \) given the original input sequence \( X \) and the preceding targets \( Y_{1:j-1} \) as predicted by the model. 
\(\Hat{X}_i\) denotes the input sequence \( X \) with the values of feature \( x_{i} \) permuted.

\subsection{Experimental Setting}
\paragraph{Data Slice}
In BlocksWorld, the dataset is randomly split into a training set (100 samples) and a validation set (500 samples). 
In TravelPlanner, we use the original training (45 samples) and validation sets (180 samples) for our experiments. 
Episodic and parametric memory updating either summarize insights from prior attempts on the training set or train on it, with evaluation performed on the validation set.
Due to the high computational cost, for BlocksWorld, we randomly select 200 samples when computing attribution scores.

\paragraph{Episodic Memory Updating}
For episodic memory updating, following previous work~\citep{zhao2024expel,fu2024autoguide}, we require language agents to summarize insights from previous attempts, categorized into two groups and one additional human-written reference:
\textit{1)} \textbf{Behavioral Cloning} --- The agent is provided with previous failed attempts along with a ground truth plan (from the external solver in BlocksWorld or human annotations in TravelPlanner).
\textit{2)} \textbf{Oracle Feedback} --- The agent is provided with previous failed attempts along with feedback from the solver or evaluator, explaining the reasons for failure.
\textit{3)} \textbf{Reference} --- This setting provides human-written insights, serving as a ground truth summary of the constraints. 
Specifically, in BlocksWorld, this refers to the reiteration of constraint descriptions, and in TravelPlanner, it includes refined summaries of commonsense and hard constraints.\footnote{TravelPlanner prohibits providing evaluation metrics directly to the agent. This setting is used solely for analysis, and the results will not be submitted to the leaderboard.}
We use \textbf{Reference} to implement the episodic memory updating strategy in the feature attribution study to avoid discrepancies in insights generated by different agents and ensure experimental control and consistency in our analysis.
More details are provided in Appendix \ref{appendix-epi-details}.

\paragraph{Parametric Memory Updating}
We use supervised fine-tuning (SFT) for parametric memory updating, with the ground truth in the training set as the optimization objective. 
All local training and inference experiments are conducted on 8$\times$A100 GPUs. 
For OpenAI models, we use the official scripts for training.
Please refer to Appendix \ref{appendix-para-details} for more details.

\section{Why Do Current Language Agents Struggle with Planning?}
\label{sec:rq1}

\begin{figure*}[t]
\centering
    \includegraphics[width=\linewidth]{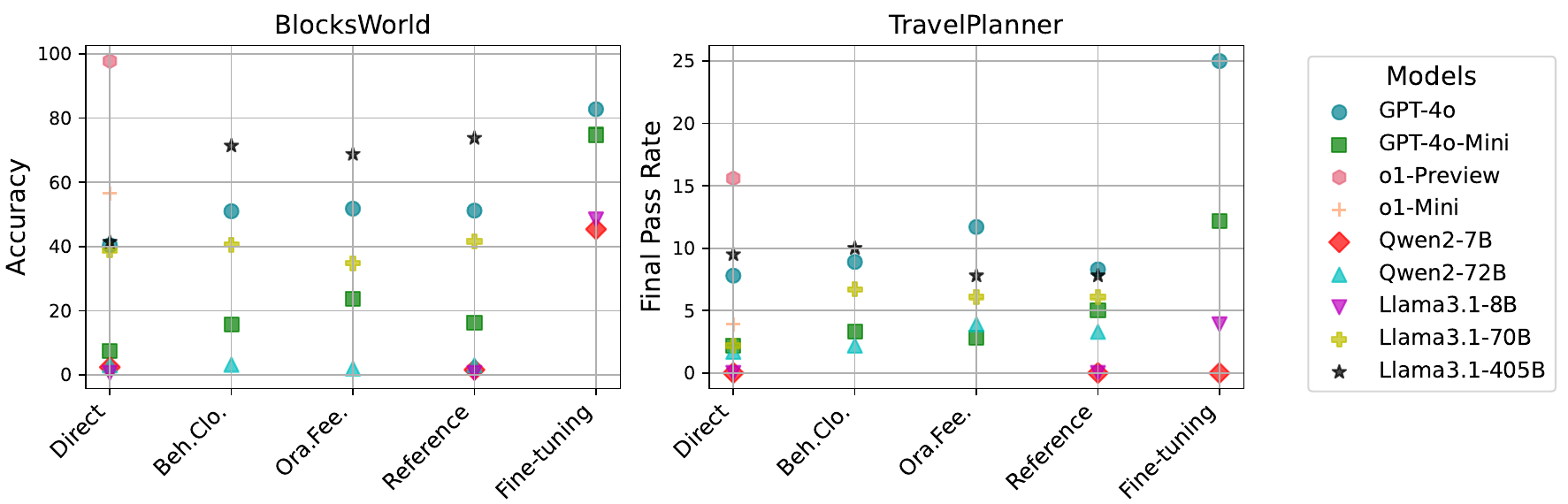}
    \caption{Main results of 9 models with different strategies on two benchmarks. 
    The results of o1-Preview and o1-Mini on BlocksWorld are from \citet{valmeekam2024llms}. ``Beh.Clo.'' and ``Ora.Fee.'' indicate Behavioral Cloning and Oracle Feedback, respectively. Llama3.1-8B and Qwen2-7B tend to provide case-specific insights that lack general applicability; thus, these models are excluded from the ``Beh.Clo.'' and ``Ora.Fee.'' settings. 
    }
    \vspace{-1em}
    \label{fig:main_res}
\end{figure*}

\subsection{Status Quo}
Agents demonstrate performance nearing or on par with humans in tasks like tool-use and web navigation~\citep{lu2023chameleon,zheng2023seeact}. 
However, when it comes to planning, which requires advanced reasoning, such as goal decomposition, constraint analysis, and foresight, agents still face significant challenges.
Specifically, as Figure~\ref{fig:main_res} demonstrates, with direct prompting, most of the current agents only complete less than half of the tasks in BlocksWorld. 
In a more complex, real-world benchmark TravelPlanner, agent performance is even lower, with none surpassing a 20\% final pass rate, including OpenAI's flagship reasoning model o1. 
This raises an important question: \textbf{Why do current language agents struggle with planning, and what hinders them from achieving advanced planning capabilities?}
In this section, we delve into the core aspects of planning to uncover the reasons behind the weak performance.

\begin{figure*}[t]
    \centering
    \includegraphics[width=\textwidth]{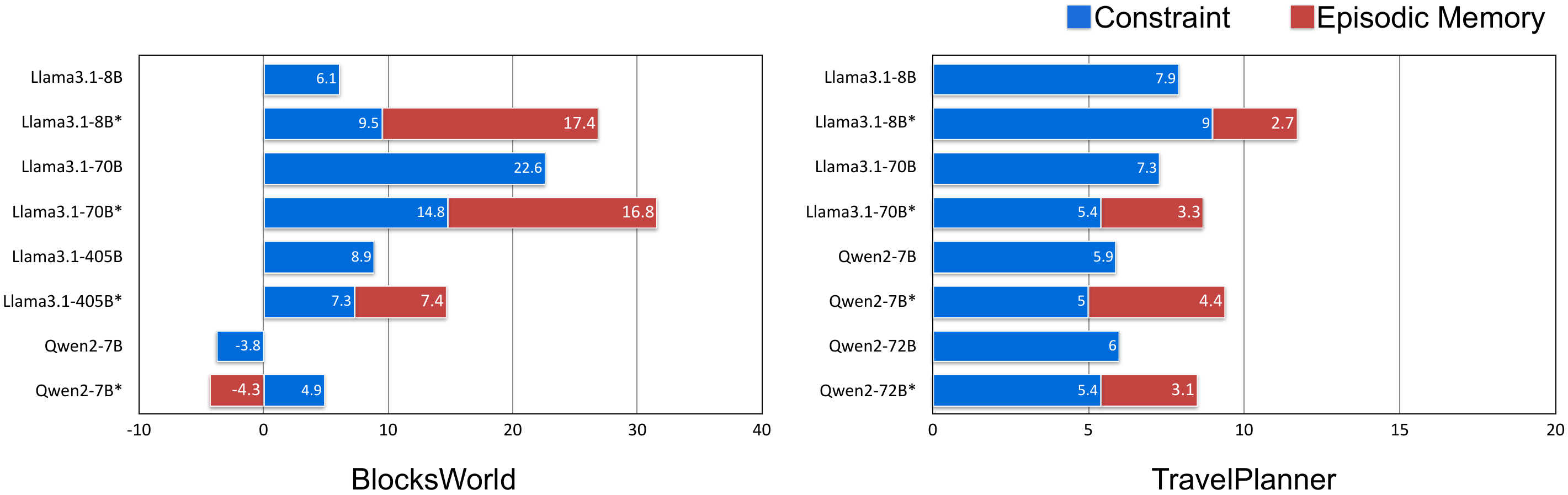}
    \caption{The attribution score of the constraint and episodic memory component in relation to the final plan across different agents, with ``*'' indicating episodic memory updating. All results are normalized to account for varying step lengths and model differences, with a maximum score of 100 representing a dominant role. The absolute value does not directly determine performance, as it only shows whether the agent references specific parts of the prompt, with factors like questions and fine-grained references also contributing. Llama3.1-405B and Qwen2-72B are selected based on performance gains from episodic memory updating and computational efficiency.}
    \label{fig:constraint_attr}
    \vspace{-1em}
\end{figure*}

\subsection{Limited Role of Constraints and Diminishing Influence of Questions}
We use Permutation Feature Importance~\citep{breiman2001random,fisher2019all} as the analysis strategy to evaluate the attribution score of each part of the prompts in relation to the final plan, covering various foundation models and two benchmarks. 

Specifically, for BlocksWorld, we divide the prompts into three components: action definitions, constraint descriptions, and questions. 
Each part is replaced with an empty token to compute its attribution score relative to the final plan.
For TravelPlanner, a real-world benchmark that requires actions to rely on commonsense embedded in the model's parameters, we focus on the attribution score of constraints and questions.
When evaluating the constraint component, we replace the attributes (e.g., price) of elements selected by the agents with an empty token to evaluate whether agents are genuinely incorporating constraints into their planning or merely generating constraint-conforming plans by chance.
For the question component, we apply the same substitution strategy used in BlocksWorld.

\paragraph{Agents do not adequately reference constraints during planning.} 
Constraints, as one of the key restraining factors, play a crucial role in planning. 
Violating constraints directly leads to failed plans since it results in illegal actions or unsatisfied goals, as highlighted in previous studies~\citep{valmeekam2024planbench,xietravelplanner}.
To investigate the reason why agents cannot obey constraints, we compute the attribution score of the constraint component, with the results presented in Figure \ref{fig:constraint_attr}. 
Compared to the upper bound score (\num{100}), which indicates a dominant role, constraints account for only a small portion of the planning process, with all scores being less than \num{25}.

\begin{figure}[t]
    \centering
    \includegraphics[width=0.48\textwidth]{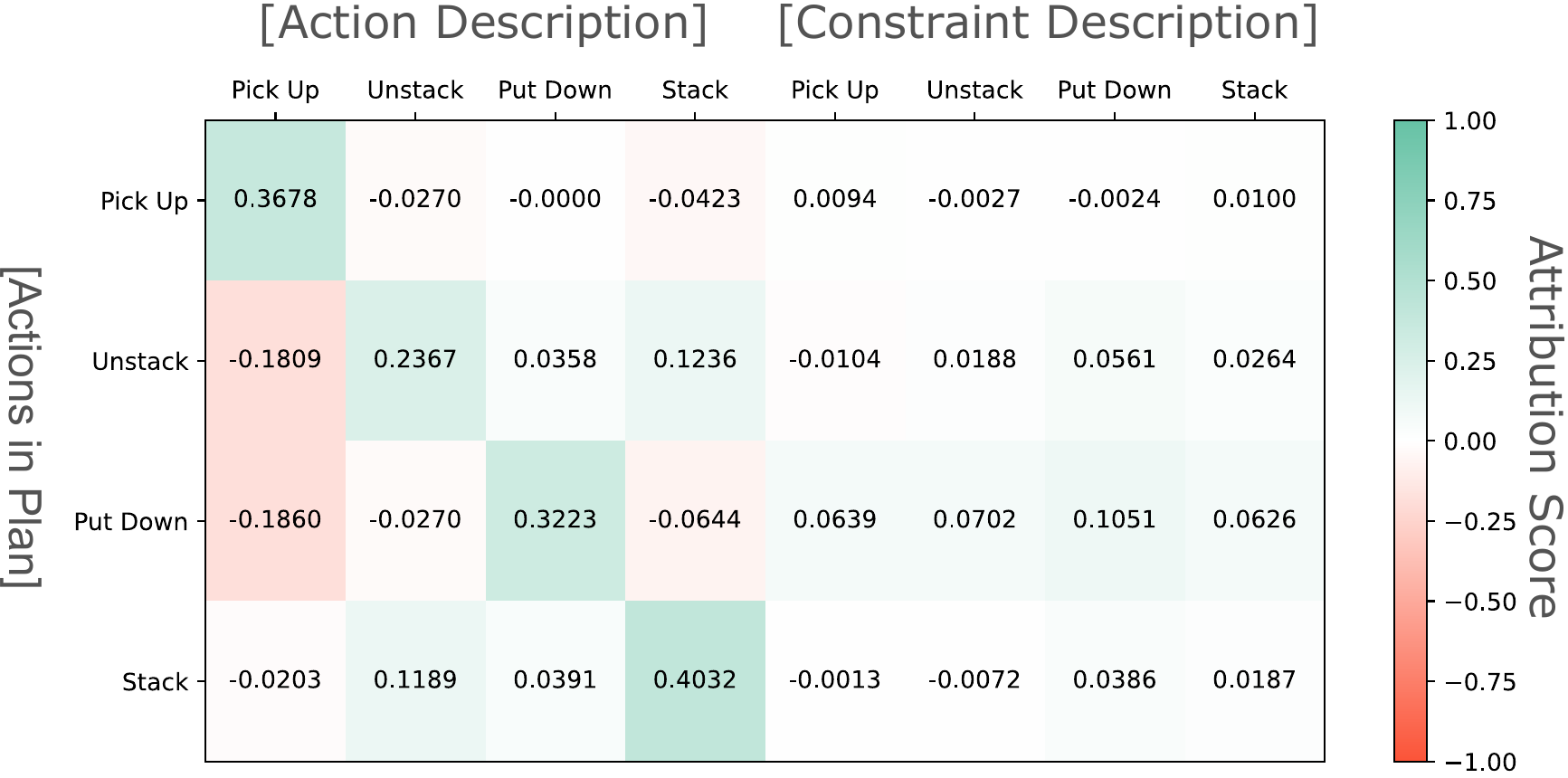}
    \caption{The distribution of attribution scores for action and constraint descriptions relative to the actions in the final plans in Llama3.1-70B on BlocksWorld. The distribution of attribution scores and discussion of TravelPlanner are in Appendix \ref{appendix-cons-disc-tp}.}
    \label{fig:llama3.1-attr-bw}
    \vspace{-1em}
\end{figure}

Furthermore, we find agents are not able to reference constraints precisely.
For example, when executing the action ``Pick Up'', agents should reference all related constraint descriptions for ``Pick Up'', and the attribution score should be significantly positive, as the constraint description contributes to the final plan. 
However, as shown by the detailed score distribution of the Llama3.1-70B model in Figure \ref{fig:llama3.1-attr-bw}, agents exhibit weak constraint-referencing behavior.
Comparing the left side of the figure, where actions align with their descriptions, the current agents fail to reference constraints effectively during planning.
Similarly, in TravelPlanner, if agents were planning based on the required attributes, the attribution score between the attribute and the final item selection should be significantly positive. 
For example, the price should influence the item choice if the agent is performing real reasoning. 
Yet, across both benchmarks, none of the agents show the ability to fully adhere to this constraint-referencing behavior, resulting in unmet preconditions in BlocksWorld and unfulfilled hard constraints in TravelPlanner.

Moreover, in BlocksWorld, we find that for Qwen2-7B, the attribution score of the constraint component is negative, indicating that the presence of constraints negatively impacts planning. 
To investigate further, we test the performance of three models---Qwen2-7B, Llama3.1-8B, and Llama3.1-70B---when no constraint descriptions are provided. 
As shown in Table \ref{tab:cons_comp}, removing constraints leads to higher scores for Qwen2-7B, while Llama3.1-70B exhibited a significant decline.
% , which also verifies the effectiveness of Permutation Feature Importance. 
This suggests that agents struggle to effectively reference constraints during planning, and in some weaker agents, constraints may even distract them and degrade their performance, which also verifies the effectiveness of our analysis strategy.

\begin{table}[t]
  \centering
    \centering
    \resizebox{\linewidth}{!}{
\begin{tabular}{lcc}
\toprule
          & \multicolumn{1}{l}{w/ Constraints} & \multicolumn{1}{l}{w/o Constraints} \\ \midrule
Qwen2-7B     & \num{2.4}                                & \num{3.6}                                \\
Llama3.1-8B  & \num{0.6}                                & \num{0.6}                                 \\

Llama3.1-70B & \num{38.8}                               & \num{9.8}                                    \\ 
Qwen2-7B$_{sft}$ & \num{45.4}                               & \num{45.4}                                    \\
Llama3.1-8B$_{sft}$ & \num{48.4}                               & \num{45.8}                                    \\
\bottomrule
\end{tabular}}
    
    \caption{Performance comparison with and without constraint descriptions in the prompts on BlocksWorld.}
    \vspace{-1em}
    \label{tab:cons_comp}
\end{table}

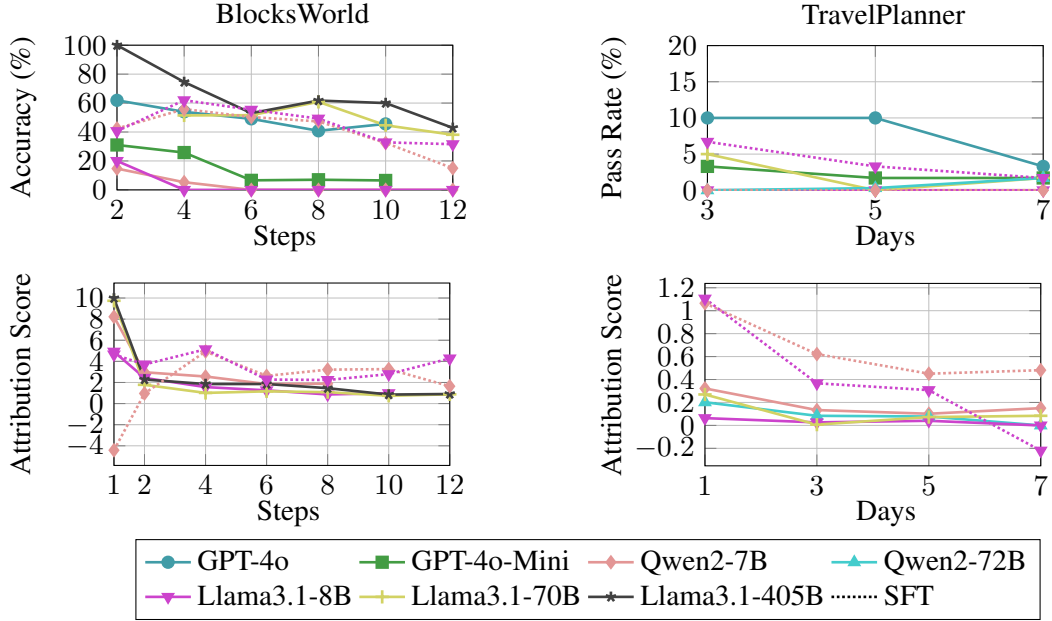
\begin{figure*}[t]
    \centering
    \begin{minipage}{.48\textwidth}
        \centerline{BlocksWorld}
        \begin{tikzpicture}
\begin{axis}[
    width=6cm, height=3.5cm,  
    xlabel={Steps},  
    ylabel={Accuracy (\%)},
    xmin=2, xmax=12,  
    ymin=0, ymax=100,  
    xtick={2,4,6,8,10,12},  
    ytick={0,20,40,60,80,100},  
    legend to name=sharedlegend,
    legend style={at={(0.5,-0.2)}, anchor=north, legend columns=4},  
    x label style={at={(axis description cs:0.5,0.1)},anchor=north},
    grid=both,  
    major grid style={line width=.2pt,draw=gray!50},  
    minor grid style={line width=.1pt,draw=gray!20}   
]

\addplot[
    color=GPT4oColor,  
    mark=*,  
    line width=1pt  
] coordinates {
(2, 61.904761904761905)
(4, 53.93258426966292)
(6, 49.101796407185624)
(8, 40.845070422535215)
(10, 45.588235294117645)
};
\addlegendentry{GPT-4o}

\addplot[
    color=GPT4oMiniColor,  
    mark=square*,  
    line width=1pt  
] coordinates {
(2, 31.03448275862069)
(4, 25.806451612903224)
(6, 6.60377358490566)
(8, 7.017543859649122)
(10, 6.521739130434782)
};
\addlegendentry{GPT-4o-Mini}

\addplot[
    color=Qwen2-7BColor,  
    mark=diamond*,  
    line width=1pt  
] coordinates {
(2, 14.705882352941178)
(4, 5.303030303030303)
(6, 0.0)
(8, 0.0)
(10, 0.0)
(12, 0.0)
};
\addlegendentry{Qwen2-7B}

\addplot[
    color=Llama3.1-8BColor,  
    mark=triangle*,  
    mark options={rotate=180},  
    line width=1pt,  
] coordinates {
    (2,20) (4,0) (6,0) (8,0) (10,0) (12,0)
};
\addlegendentry{Llama3.1-8B}

\addplot[
    color=Llama3.1-70BColor,  
    mark=x,  
    line width=1pt,  
    mark options={solid,rotate=45},
    mark size=2.5pt
] coordinates {
(4, 51.470588235294116)
(6, 51.61290322580645)
(8, 60.83333333333333)
(10, 44.642857142857146)
(12, 38.095238095238095)
};
\addlegendentry{Llama3.1-70B}

\addplot[
    color=Llama3.1-405BColor,  
    mark=star,  
    line width=1pt  
] coordinates {
(2, 100.0)
(4, 74.54545454545455)
(6, 53.04347826086957)
(8, 61.73913043478261)
(10, 60.0)
(12, 42.857142857142854)
};
\addlegendentry{Llama3.1-405B}

\addplot[
    color=Qwen2-7BColor,  
    mark=diamond*,  
    line width=1pt,  
    dash pattern=on 1pt off 1pt,  
    mark options={solid},  
] coordinates {
(2, 42.42424242424242)
(4, 55.81395348837209)
(6, 50.33557046979866)
(8, 47.32824427480916)
(10, 32.432432432432435)
(12, 15.0)
};
\addlegendentry{Qwen2-7B$_{sft}$}

\addplot[
    color=Llama3.1-8BColor,  
    mark=triangle*,  
    line width=1pt,
    mark options={rotate=180,solid},
    dash pattern=on 1pt off 1pt,  
] coordinates {
(2, 40.625)
(4, 61.95652173913043)
(6, 55.39568345323741)
(8, 49.25373134328358)
(10, 32.83582089552239)
(12, 31.57894736842105)
};
\addlegendentry{Llama3.1-8B$_{sft}$}

\end{axis}
\end{tikzpicture}      
        \label{fig:constraint_attr_bw}

    \end{minipage}
    \begin{minipage}{.48\textwidth}
        \centerline{TravelPlanner}
        \begin{tikzpicture}
\begin{axis}[
    width=6cm, height=3.5cm,  
    xlabel={Days},  
    ylabel={Pass Rate (\%)},  
    xmin=3, xmax=7,  
    ymin=0, ymax=20,  
    xtick={3,5,7},  
    ytick={0,5,10,15,20},  
    grid=both,  
    legend to name=sharedlegend,
    legend style={at={(0.5,-0.2)}, anchor=north, legend columns=4},  
    grid=both,  
    x label style={at={(axis description cs:0.53,0.1)},anchor=north},
    major grid style={line width=.2pt,draw=gray!50},  
    minor grid style={line width=.1pt,draw=gray!20}   
]

\addplot[
    color=GPT4oColor,  
    mark=*,  
    line width=1pt  
] coordinates {
    (3,10) (5,10) (7,3.3)
};
\addlegendentry{GPT-4o}

\addplot[
    color=GPT4oMiniColor,  
    mark=square*,  
    line width=1pt  
] coordinates {
    (3,3.3) (5,1.7) (7,1.7)
};
\addlegendentry{GPT-4o-Mini}

\addplot[
    color=Qwen2-7BColor,  
    mark=diamond*,  
    line width=1pt  
] coordinates {
    (3,0) (5,0) (7,0)
};
\addlegendentry{Qwen2-7B}

\addplot[
    color=Llama3.1-8BColor,  
    mark=triangle*, mark options={rotate=180},  
    line width=1pt  
] coordinates {
    (3,0) (5,0) (7,0)
};
\addlegendentry{Llama3.1-8B}

\addplot[
    color=Llama3.1-70BColor,  
    mark=x,  
    line width=1pt,  
    mark options={solid,rotate=45},
    mark size=2.5pt
] coordinates {
    (3,5) (5,0) (7,1.7)
};
\addlegendentry{Llama3.1-70B}

\addplot[
    color=Qwen2-72BColor,  
    mark=triangle*,  
    line width=1pt  
] coordinates {
    (3,0) (5,0.3) (7,1.7)
};
\addlegendentry{Qwen2-72B}

\addplot[
    color=Qwen2-7BColor,  
    mark=diamond*,  
    line width=1pt,  
    dash pattern=on 1pt off 1pt,  
    mark options={solid}
] coordinates {
    (3,0) (5,0) (7,0)
};
\addlegendentry{Qwen2-7B$_{sft}$}

\addplot[
    color=Llama3.1-8BColor,  
    mark=triangle*,  
    mark options={rotate=180,solid},
    line width=1pt,  
    dash pattern=on 1pt off 1pt,  
] coordinates {
    (3,6.7) (5,3.3) (7,1.7)
};
\addlegendentry{Llama3.1-8B$_{sft}$}
\end{axis}
\end{tikzpicture}      
        \label{fig:}
    \end{minipage}

    \begin{minipage}{0.48\textwidth}
        \begin{tikzpicture}
\begin{axis}[
    width=6cm, height=4cm, 
    xlabel={Steps},
    ylabel={Attribution Score},  
    xmin=1, xmax=12, 
    xtick={1,2,4,6,8,10,12}, 
    ytick={-4,-2,0,2,...,10},  
    grid=both, 
    major grid style={line width=.2pt,draw=gray!50}, 
    minor grid style={line width=.1pt,draw=gray!20},  
    legend to name=sharedlegend,
    x label style={at={(axis description cs:0.52,0.08)},anchor=north},
    legend style={at={(0.5,-0.2)}, anchor=north, legend columns=4},  
]

\addplot[
    color=GPT4oColor, 
    mark=*,  
    line width=1pt  
] coordinates {
(0,0)
};
\addlegendentry{GPT-4o}

\addplot[
    color=GPT4oMiniColor,  
    mark=square*,  
    line width=1pt 
] coordinates {
(0,0)
};
\addlegendentry{GPT-4o-Mini}

% 第二条折线
\addplot[
    color=Qwen2-7BColor,  
    mark=diamond*, 
    line width=1pt  
] coordinates {
( 1 , 8.222352027893066 )
( 2 , 2.96646785736084 )
( 4 , 2.565281391143799 )
( 6 , 1.8504897356033325 )
( 8 , 1.880358099937439 )
};

% 添加图例
\addlegendentry{Qwen2-7B}

% 绘制折线图
\addplot[
    color=Llama3.1-8BColor,
    mark=triangle*, 
    mark options={rotate=180}, 
    line width=1pt 
] coordinates {
( 1 , 4.91966438293457 )
( 2 , 2.434152126312256 )
( 4 , 1.5616358518600464 )
( 6 , 1.2544955015182495 )
( 8 , 0.8649712204933167 )
( 10 , 0.9873016476631165 )
};

\addlegendentry{Llama3.1-8B}

% 第二条折线
\addplot[
    color=Llama3.1-70BColor, 
    mark=x, 
    line width=1pt, 
    mark options={solid,rotate=45},
    mark size=2.5pt
] coordinates {
( 1 , 9.73923397064209 )
( 2 , 1.7803089618682861 )
( 4 , 1.011332631111145 )
( 6 , 1.1592504978179932 )
( 8 , 1.0821706056594849 )
( 10 , 0.7410396933555603 )
( 12 , 0.8774887323379517 )
};

% 添加图例
\addlegendentry{Llama3.1-70B}

\addplot[
    color=Llama3.1-405BColor,  
    mark=star,  
    line width=1pt  
] coordinates {
( 1 , 9.979561805725098 )
( 2 , 2.2805447578430176 )
( 4 , 1.8521219491958618 )
( 6 , 1.854600191116333 )
( 8 , 1.4594439268112183 )
( 10 , 0.8545204997062683 )
( 12 , 0.9078516960144043 )
};
\addlegendentry{Llama3.1-405B}

\addplot[
    color=Qwen2-7BColor,  
    mark=diamond*,  
    line width=1pt,  
    mark options={solid},
    dash pattern=on 1pt off 1pt, 
] coordinates {
( 1 , -4.418802261352539 )
( 2 , 0.9600092172622681 )
( 4 , 4.936853408813477 )
( 6 , 2.626099109649658 )
( 8 , 3.2162764072418213 )
( 10 , 3.264580249786377 )
( 12 , 1.6423472166061401 )
};
\addlegendentry{Qwen2-7B$_{sft}$}

\addplot[
    color=Llama3.1-8BColor,  
    mark=triangle*, 
    line width=1pt,
    mark options={rotate=180,solid},
    dash pattern=on 1pt off 1pt, 
] coordinates {
( 1 , 4.6474103927612305 )
( 2 , 3.7084288597106934 )
( 4 , 5.143255710601807 )
( 6 , 2.2771947383880615 )
( 8 , 2.2365801334381104 )
( 10 , 2.7865734100341797 )
( 12 , 4.274865627288818 )
};
\addlegendentry{Llama3.1-8B$_{sft}$}

\end{axis}
\end{tikzpicture}
        \label{fig:question_attr_trend_bw}
    \end{minipage}
    \begin{minipage}{0.48\textwidth}
        \begin{tikzpicture}
\begin{axis}[
    width=6cm, height=4cm,  
    xlabel={Days},  
    ylabel={Attribution Score},  
    xmin=1, xmax=7,  
    xtick={1,3,5,7},  
    ytick={-0.2,0,0.2,...,1.2},  
    grid=both,  
    major grid style={line width=.2pt,draw=gray!50},  
    minor grid style={line width=.1pt,draw=gray!20},  
    legend to name=sharedlegend,
    x label style={at={(axis description cs:0.54,0.08)},anchor=north},
    legend style={at={(0.5,-0.2)}, anchor=north, legend columns=4, legend cell align=left},  
]

\addplot[
    color=GPT4oColor,  
    mark=*,  
    line width=1pt  
] coordinates {
(0,0)
};
\addlegendentry{GPT-4o}

\addplot[
    color=GPT4oMiniColor,  
    mark=square*,  
    line width=1pt  
] coordinates {
(0,0)
};
\addlegendentry{GPT-4o-Mini}

\addplot[
    color=Qwen2-7BColor,  
    mark=diamond*,  
    line width=1pt  
] coordinates {
(1, 0.3223305344581604)
(3, 0.13357481360435486)
(5, 0.10148869454860687)
(7, 0.15033581852912903)
};

\addlegendentry{Qwen2-7B}

\addplot[
    color=Qwen2-72BColor,  
    mark=triangle*,  
    line width=1pt  
] coordinates {
(1, 0.20152153074741364)
(3, 0.08358651399612427)
(5, 0.0774524137377739)
(7, -0.00048302410868927836)
};

\addlegendentry{Qwen2-72B}

\addplot[
    color=Llama3.1-8BColor,  
    mark=triangle*, 
    mark options={rotate=180},  
    line width=1pt  
] coordinates {
(1, 0.0633578822016716)
(3, 0.02619488537311554)
(5, 0.040252309292554855)
(7, -0.0008663424523547292)
};

\addlegendentry{Llama3.1-8B}

\addplot[
    color=Llama3.1-70BColor,  
    mark=x,  
    line width=1pt,  
    mark options={solid,rotate=45},
    mark size=2.5pt
] coordinates {
(1, 0.26961684226989746)
(3, 0.006657848134636879)
(5, 0.07251884043216705)
(7, 0.08365966379642487)
};

\addlegendentry{Llama3.1-70B}

\addplot[
    color=Llama3.1-405BColor,  
    mark=star,  
    line width=1pt
] coordinates {
(0, 0)
};
\addlegendentry{Llama3.1-405B}

\addplot[
    color=Qwen2-7BColor,  
    mark=diamond*,  
    line width=1pt,  
    dash pattern=on 1pt off 1pt, 
    mark options={solid},
    forget plot
] coordinates {
(1, 1.0647386312484741)
(3, 0.6216356158256531)
(5, 0.45064035058021545)
(7, 0.4812552332878113)
};

\addplot[
    color=Llama3.1-8BColor,  
    mark=triangle*,  
    line width=1pt,  
    dash pattern=on 1pt off 1pt, 
    mark=triangle*, 
    mark options={solid,rotate=180},
    forget plot  
] coordinates {
(1, 1.1052637100219727)
(3, 0.3672389090061188)
(5, 0.30751755833625793)
(7, -0.22000545263290405)
};

\addplot[
    color=black,  
    line width=1pt,  
    dash pattern=on 1pt off 1pt, 
    mark options={solid,rotate=180}
] coordinates {
(0, 0)
};
\addlegendimage{color=black, mark=*, line width=1pt, dash pattern=on 1pt off 1pt}
\addlegendentry{SFT}
\end{axis}
\end{tikzpicture}
        \label{fig:question_attr_trend_tp}
    \end{minipage}
    \pgfplotslegendfromname{sharedlegend}
    \caption{Performance comparison with increasing planning horizon. The upper part shows the performance of different agents, while the lower part shows their attribution scores of questions as the planning horizon extends. }
    \label{fig:step_comp}
\end{figure*}

\paragraph{As the planning horizon increases, the influence of the question on plan generation decreases.}
We find that while agents can deliver a complete plan, these plans often fail to meet the goals specified in the question, with failure rates increasing as the planning horizon increases. 
Specifically, as shown in the upper part of Figure \ref{fig:step_comp}, in both BlocksWorld and TravelPlanner, agent performance declines as the number of generated steps or travel days increases. 
Could this be similar to the ``lost in the middle'' phenomenon~\citep{liu-etal-2024-lost}, where they lose track of the goal as the planning horizon increases?
To investigate the underlying reasons, we compute the attribution score of the question at different steps in the plan, as shown in the lower part of Figure \ref{fig:step_comp}. 

We find that as the planning horizon increases, the attribution score of the question decreases. 
This suggests that the question’s influence on specific action or item selection diminishes as the plan progresses. 
This explains why agents perform worse as the plan length increases: if agents fail to focus on the goal specified in the query and lose track of it, they cannot deliver a successful plan.

\section{What Happens in Memory Updating for Language Agents?}
\label{sec:rq2}

While previous work and our experiments, as shown in Figure~\ref{fig:main_res}, show that both parametric memory updating and episodic memory updating can improve agents' performance in planning tasks, the underlying mechanisms remain unclear. 
In this section, we aim to address the following two questions:
\textit{1)} Why do memory updating strategies help improve agents' planning abilities?
\textit{2)} What limitations of these strategies prevent agents from achieving more advanced planning abilities?

\begin{figure*}[t]
    \centering
    \begin{minipage}{0.45\textwidth}
        \centering  
        \includegraphics[width=\textwidth]{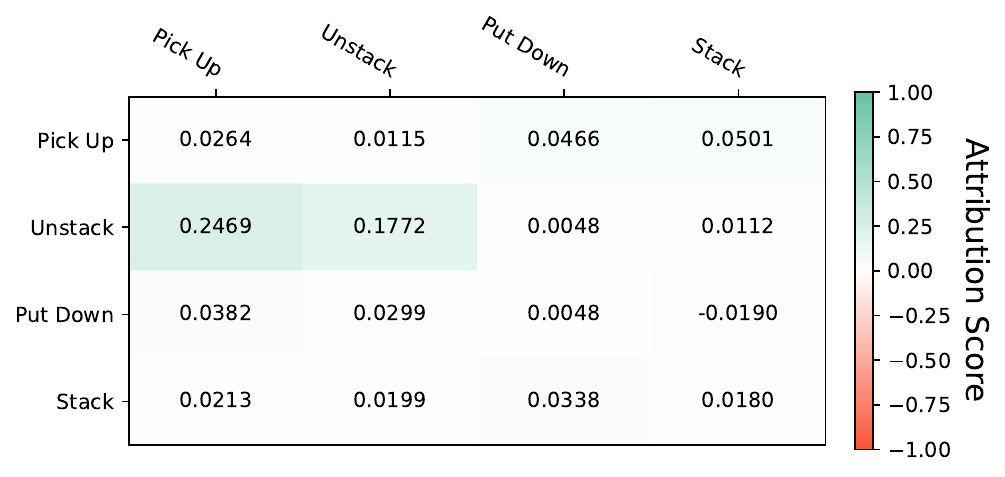}
        \centerline{BlocksWorld}
        \label{fig:llama3.1-attr-70B_insight_bw}
    \end{minipage}%
    \begin{minipage}{0.47\textwidth}
        \centering  
        \includegraphics[width=\textwidth]{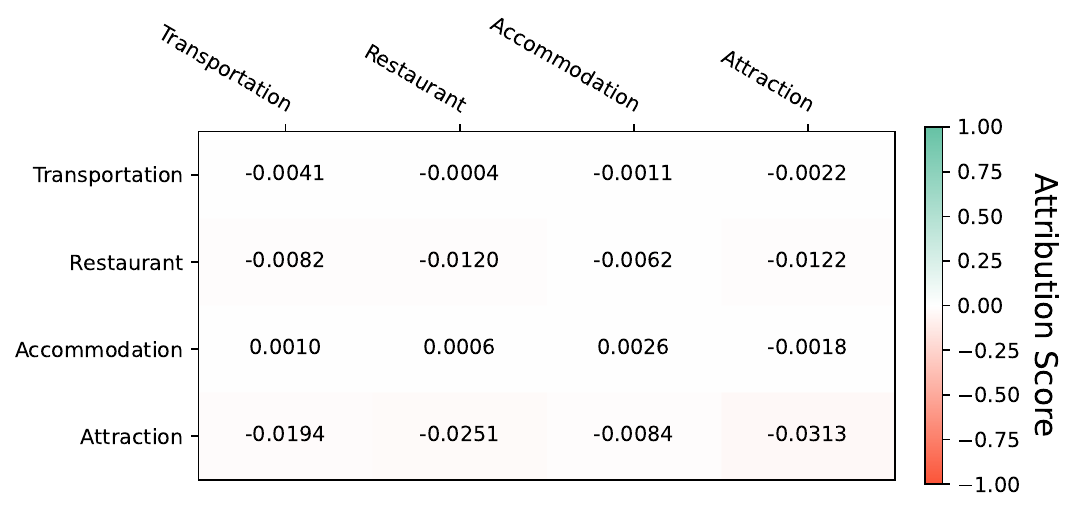}
        \centerline{TravelPlanner}
        \label{fig:constraint_attr_tp}
    \end{minipage}
    \vspace{-1.5em}
    \caption{The attribution scores of episodic memory to the final plan in Llama3.1-70B on two benchmarks. The abscissa is the constraint, and the ordinate is the corresponding action or item in the plan. }
    \vspace{-1em}
    \label{fig:ind_constraint_attr}
\end{figure*}

\subsection{Episodic memory Updating }

\paragraph{Episodic memory updating refines and reiterates constraint information, making it easier for agents to recognize and apply.}

Rather than incorporating new information, we find that simply refining or reiterating existing insights in episodic memory updating can lead to performance improvements. 
In TravelPlanner, performance gains are observed when refined information (e.g., insights like selecting cheaper items, which agents would otherwise need to infer themselves) is introduced. 
Similarly, in BlocksWorld, both agent-generated and human-written insights---despite being slight modifications or emphases of the original constraint descriptions---still result in performance enhancements with episodic memory updating. 
This is intriguing, as such repetition typically offers little value in human reasoning.
 
To assess the impact of episodic memory updating on plan generation, we compute the attribution score of episodic memory (Figure \ref{fig:constraint_attr}). 
Specifically, these refined or reiterated insights show positive attribution scores to the final plans, indicating that agents actively consider them during planning. 
However, the figure also shows that while vague and implicit episodic memories (e.g., ``select cheap items'' in TravelPlanner) do contribute, more explicit and direct constraints (e.g., ``cannot 'Pick Up' when the hand is not empty'' in BlockWorld) are easier for agents to utilize, as demonstrated by their higher attribution scores.

\paragraph{Agents understand episodic memory on a global level and cannot reference it in a fine-grained manner.}
While episodic memory updating improves performance, the gains remain relatively minor. 
To investigate this further, we decompose the episodic memory into discrete components (i.e., treating each insight independently) to assess whether agents can reference these insights in a fine-grained manner. 
For example, in TravelPlanner, agents are expected to consider specific insights related to accommodation when selecting lodging options.
However, as shown in Figure \ref{fig:ind_constraint_attr}, while agents reference the overall episodic memory during planning, they struggle to apply individual insights in a detailed, fine-grained manner, reflected in their relatively low scores.

Moreover, in BlocksWorld, we observe that the constraint description for ``Unstack'' plays only a minor role (scoring \num{0.0188}) in the original constraint attribution (Figure~\ref{fig:llama3.1-attr-bw}), but contributes significantly more in the episodic memory (\num{0.1954}; Figure~\ref{fig:ind_constraint_attr}). 
We hypothesize that episodic memory complements information that agents might have initially overlooked. 
However, agents still struggle to apply this information effectively during planning when dealing with more vague or implicit episodic memories, such as those in TravelPlanner.

\subsection{Parametric Memory Updating}

\paragraph{Parametric memory updating improves the attribution score of questions.}
Although parametric memory updating improves agents' performance in planning tasks, its underlying mechanisms remain unclear. 
Building on previous findings that the attribution score of questions relates to an agent's planning performance, we investigate whether this score changes after parametric memory updating. 
As shown in Figure \ref{fig:step_comp}, we observe a positive correlation between the question attribution score and final performance. 
For example, in BlocksWorld, from step 2 to step 4, the question attribution score increases, resulting in both fine-tuned agents achieving their highest scores at step 4.
This suggests that through fine-tuning, agents are able to have a stronger focus on the goal than before, leading to improved planning outcomes.

\paragraph{While parametric memory updating increases the attribution score of questions, it still struggles as the planning horizon increases.}
Despite improvements in question attribution scores after fine-tuning compared to the vanilla agents, a decline is observed after step 4 in BlocksWorld, leading to a corresponding drop in performance. 
A similar trend is also noted in TravelPlanner.
This suggests a promising direction:  maintaining a strong focus on the goal throughout planning is essential for overcoming short-horizon limitations and advancing agents' planning abilities.

\begin{table}[]
\centering
\begin{tabular}{lcc}
\toprule
Episodic Memory              & \multicolumn{1}{c}{$\times$} & \multicolumn{1}{c}{\checkmark} \\ \midrule
Qwen2-7B$_{sft}$     & \num{45.4}                                & \num{43.0}                              \\

Llama3.1-8B$_{sft}$ & \num{48.4}                              & \num{36.8}                                   \\ 
\bottomrule
\end{tabular}
\caption{Comparison between two fine-tuned models with and without episodic memory updating on BlocksWorld.}
\vspace{-1em}
\label{tab:ft_model_with_insight}
\end{table}

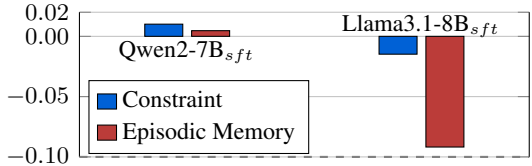
\begin{figure}[t]
    \centering
    \definecolor{ConstraintColor}{RGB}{0, 97, 214}  
\definecolor{EpisodicMemoryColor}{RGB}{187, 60, 57}   

\pgfplotsset{
    every axis legend/.append style={
        legend image code/.code={
            \draw[#1,fill=#1] (0cm,-0.1cm) rectangle (0.25cm,0.1cm);
        }
    }
}

\begin{tikzpicture}

\begin{axis}[
    width=7.5cm, height=3.5cm, 
    ymin=-0.10, ymax=0.02, 
    axis x line=none, 
    ytick={-0.10, -0.05, 0, 0.02},  
    y tick label style={
        /pgf/number format/fixed,  
        /pgf/number format/precision=2,
        /pgf/number format/fixed zerofill, 
        font=\small
    },
    ylabel style={font=\footnotesize},
    tick label style={font=\small}, 
    every node near coord/.append style={
        font=\scriptsize,  
        /pgf/number format/fixed, 
        /pgf/number format/precision=2, 
        /pgf/number format/fixed zerofill
    },  
    enlarge x limits=0.3,  
    bar width=0.5cm,  
    grid=both, 
    major grid style={line width=.2pt,draw=gray!50},
    minor grid style={line width=.1pt,draw=gray!20},
    ymajorgrids,        
    y axis line style={-}, 
    legend style={
        at={(0.02,0.02)},  
        anchor=south west,
        font=\footnotesize, 
        cells={anchor=west}  
    },
    after end axis/.code={
        \draw[thick, dashed, gray] 
        (rel axis cs:0,0) -- (rel axis cs:1,0); 
    },
]

\addplot[
    ybar,
    bar width=0.5cm,
    fill=ConstraintColor,
] coordinates {
    (1, 0.0101)
    (1.5, -0.0148)
};

\node at (axis cs:1, 0.012) [above, font=\small, xshift=0.3cm,yshift=-0.68cm] {Qwen2-7B$_{sft}$};
\node at (axis cs:1.5, 0.012) [above, font=\small, xshift=0.3cm,yshift=-0.34cm] {Llama3.1-8B$_{sft}$};

\addplot[
    ybar,
    bar width=0.5cm,
    fill=EpisodicMemoryColor,
] coordinates {
    (1.1, 0.0047)
    (1.6, -0.0918)
};

\legend{Constraint, Episodic Memory}

\end{axis}
\end{tikzpicture}
    \vspace{-1em}
    \caption{Attribution scores of constraints and episodic memory on BlocksWorld for two fine-tuned agents.}
    \label{fig:insight_attr_ft_bw}
\end{figure}

\begin{table}[t]
    \centering
    
\resizebox{\linewidth}{!}
{\begin{tabular}{cccccc}
\toprule
     
              & \multicolumn{2}{c}{Commonsense} & \multicolumn{2}{c}{Hard} & \multirow{2}{*}{\begin{tabular}[c]{@{}c@{}}Final\\ Pass Rate\end{tabular}} \\ \cmidrule(l){2-3} \cmidrule(l){4-5}
              & Micro          & Macro          & Micro       & Macro      &  
\\ \midrule

\rowcolor[gray]{0.85}
\multicolumn{6}{c}{Direct Prompting} 

\\ \midrule
       
GPT-4o        & \num{84.7}           & \num{31.1}           & \num{53.6}        & \num{31.1}       & \num{7.8}                                                                        \\
GPT-4o-Mini   & \num{84.4}           & \num{22.2}           & \num{42.4}        & \num{20.0}       & \num{2.2}                                                                        \\
                                                            
Llama3.1-8B   & \num{60.1}           & \num{0.0}              & \num{7.9}         & \num{2.8}        & \num{0.0}                                                                          \\
Llama3.1-70B  & \num{82.8}          & \num{18.9}           & \num{33.1}        & \num{16.1}       & \num{2.2}                                                                        \\

Qwen2-7B & \num{49.9}           & \num{1.1}           & \num{2.1}        & \num{0.0}      & \num{0.0} \\
Qwen2-72B & \num{74.8}           & \num{11.7}           & \num{23.8}       & \num{8.9}       & \num{1.7}    \\ \midrule

\rowcolor[gray]{0.85}
\multicolumn{6}{c}{Episodic Memory Updating}                                                                                                            \\ \midrule
GPT-4o        & \num{89.2}           & \num{41.7}           & \num{51.7}        & \num{27.2}       & \num{8.3}                                                                        \\
GPT-4o-Mini   & \num{84.1}           & \num{22.2}           & \num{39.8}        & \num{22.8}       & \num{5.0}                                                                        \\
                                                              
Llama3.1-70B  & \num{84.9}           & \num{23.9}           & \num{39.5}        & \num{24.4}       & \num{6.1}                                                                        \\

Qwen2-72B & \num{75.6}           & \num{13.8}           & \num{28.8}        & \num{10.6}       & \num{3.3} \\  
\hdashline
$\Delta$   &{\textcolor{caribbeangreen}{+\num{1.8}}}           & {\textcolor{caribbeangreen}{+\num{4.4}}}           & {\textcolor{caribbeangreen}{+\num{1.7}}}        & {\textcolor{caribbeangreen}{+\num{2.3}}}       & {\textcolor{caribbeangreen}{+\num{2.2}}}    \\
\midrule

\rowcolor[gray]{0.85}
\multicolumn{6}{c}{Parametric Memory Updating}                                                                                                          \\ \midrule
GPT-4o        & \num{95.3}           & \num{68.9}           & \num{62.6}        & \num{39.4}       & \num{25.0}                                                                       \\
GPT-4o-Mini   & \num{94.7}           & \num{61.7}           & \num{49.3}        & \num{17.2}       & \num{12.2}                                                                       \\
Llama3.1-8B   & \num{78.3}           & \num{17.8}           & \num{19.3}        & \num{6.1}        & \num{3.8}                                                                        \\ 
Qwen2-7B   & \num{59.0}           & \num{0.6}           & \num{0.2}        & \num{0.0}        & \num{0.0}    \\ 
\hdashline
$\Delta$   & {\textcolor{caribbeangreen}{+\num{12.1}}}           & {\textcolor{caribbeangreen}{+\num{23.7}}}           & {\textcolor{caribbeangreen}{+\num{6.4}}}        & {\textcolor{caribbeangreen}{+\num{2.2}}}         & {\textcolor{caribbeangreen}{+\num{7.8}}}    \\ 
\bottomrule
\end{tabular}}
    \vspace{-0.5em}
    \caption{Comparison between different agents on TravelPlanner. ``$\Delta$'' represents the average improvement compared to the same model using direct prompting.}
    \vspace{-1em}
    \label{tab:score_tp}
\end{table}

\section{Discussion}
\label{sec:discussion}

\paragraph{When constraints are already parameterized, episodic memory updating does not improve performance and may even degrade it.}

If both parametric and episodic memory updating are effective for agent planning, an interesting question arises: Would it be better to combine these two strategies? 
Surprisingly, this mixture does not improve the performance of fine-tuned agents and even harms it. 
As shown in Table \ref{tab:ft_model_with_insight}, both fine-tuned agents exhibit a performance decline after episodic memory updating.
Moreover, as shown in Figure \ref{fig:insight_attr_ft_bw}, the attribution scores of both constraints and episodic memory play only a minor or even negative role, indicating that the fine-tuned agents no longer explicitly reference these constraints, rendering them redundant and ineffective.
We hypothesize that reiterated episodic memory becomes redundant when constraints are already embedded within the model’s parameters. 
This redundancy disrupts the model's decision-making coherence and undermines its ability to leverage the pre-existing constraint knowledge, resulting in weaker planning performance.

To explore this further, we also report the performance of fine-tuned agents with the constraints removed in Table \ref{tab:cons_comp}. 
Unlike the vanilla Llama3.1-70B, which shows a noticeable performance drop, the Llama3.1-8B$_{sft}$ shows only a slight decline, and Qwen2-7B$_{sft}$ even exhibits no decrease at all, suggesting that the constraints had already been parameterized within them.

\paragraph{Both strategies resemble shortcut learning, focusing on short-horizon and low-level planning.}
Although both strategies offer performance improvements, they fail to achieve our expected high-level intelligence. 
Our findings suggest that these strategies resemble ``shortcut learning,'' favoring static rule learning over dynamic problem-solving. 
For example, in TravelPlanner, agents learn commonsense rules effectively, especially through parametric memory updating (see Table \ref{tab:score_tp}), as commonsense is often based on static patterns learned in training data. 
However, these strategies remain insufficient when faced with hard constraints requiring advanced model abilities, such as maintaining a strong focus on long-horizon tasks, precise referencing for multiple-constraint integration, and sophisticated planning skills like foresight, simulation, and backtracking for trajectory adjustments.

\section{Conclusion}
\label{sec:conclusion}
This paper utilizes Permutation Feature Importance to investigate why current language agents struggle with planning tasks. 
Our findings show that constraints play only a minor role in agent planning, indicating that agents are not effectively considering constraints during planning. 
Additionally, the question's influence diminishes as the planning horizon extends, causing agents to lose focus on the goal and resulting in failed plans.
Furthermore, we examine the effects of episodic and parametric memory updating on agent performance. 
While both strategies improve the impact of constraints and questions in planning, they only mitigate the underlying issues rather than fully resolve them.

We hope this paper provides valuable insights and sparks future research to address language agents' key challenges in planning, ultimately moving closer to achieving human-level intelligence.

\section*{Limitations}
\label{sec:limitation}
In this paper, we use Permutation Feature Importance to calculate the attribution scores across various open-source model families and sizes, aiming to provide insights into the obstacles that current language agents face in planning tasks.
We also test the performance of the widely used GPT family. 
However, due to the limited access provided by OpenAI's API, which restricts control over output token generation, we are unable to compute attribution scores for these models. 
Nonetheless, the consistent conclusions drawn from the two used model families across different sizes and benchmarks support the robustness of our methodology and validate our overall findings.

\bibliography{custom}

\clearpage

\appendix

\section*{Appendix}
\label{sec:appendix}
\setcounter{table}{0}
\renewcommand\thetable{\Alph{section}.\arabic{table}}
\setcounter{figure}{0}
\renewcommand\thefigure{\Alph{section}.\arabic{figure}}

Within this supplementary material, we elaborate on the following aspects:
\begin{itemize}
\vspace{0.5em}
\item Appendix \ref{appendix:disc}: Discussions
\vspace{0.5em}
\item Appendix \ref{appendix:experiment}: Experimental Setup Details
\vspace{0.5em}
\item Appendix \ref{appendix:prompt-list}: Prompts List
\vspace{0.3em}
\end{itemize}

\section{Discussion}
\label{appendix:disc}
\subsection{Attribution Scores of Constraint Tokens on TravelPlanner}
\label{appendix-cons-disc-tp}

\begin{figure}[h]
    \centering
    \includegraphics[width=0.48\textwidth]{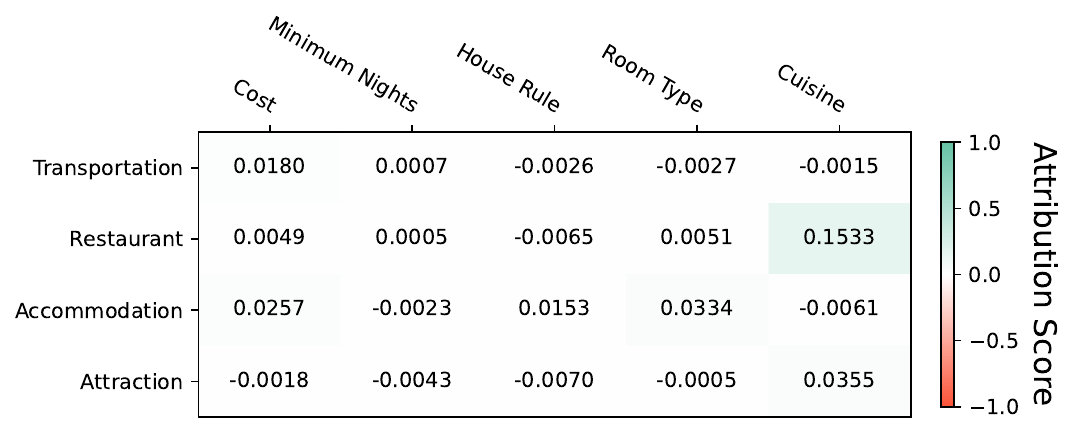}
    \caption{The distribution of attribution scores for constraint descriptions relative to the actions in the final plan in Llama3.1-70B on TravelPlanner.}
    \vspace{-1em}
    \label{fig:appendix-llama3.1-cons-attr-tp}
\end{figure}

As shown in Figure \ref{fig:appendix-llama3.1-cons-attr-tp}, aside from the cuisine attribute in the prompts, other item attributes contribute minimally to the final plan. 
This suggests why the agent struggles to follow commonsense and hard constraints in TravelPlanner, as the key attributes that determine whether constraints are followed or violated play only minor roles. 
This further highlights that current agents still face challenges in fully integrating multiple attributes and adhering to constraints effectively when generating satisfactory plans.

\section{Experimental Setup Details}
\label{appendix:experiment}
\subsection{Episodic Memory Updating}
\label{appendix-epi-details}

\paragraph{Training}

For episodic memory updating, we follow the methodology proposed by \citet{zhao2024expel}, where the agent is tasked with summarizing insights from previous attempts. The process of filtering these insights involves a voting system, where the agent can take one of the following actions:

\begin{itemize}
    \item \textbf{Add:} Introduce new, general insights that are not restricted to specific queries and are missing from the current set. New insights are beginning with one vote.
    \item \textbf{Modify:} Revise existing insights if they are incomplete or partially incorrect. This action preserves the original number of votes for the insight.
    \item \textbf{Support:} Endorse correct insights by increasing their vote count by one. This action ensures that useful insights are retained and emphasized. 
    \item \textbf{Oppose:} Challenge incorrect or irrelevant insights, decreasing their vote count by one. This process helps eliminate inaccuracies.
\end{itemize}

\paragraph{Inference}

When inference in the validation set, agents are required to use the insight learned in the training set. 
First, they are required to select the insight that they think is useful and then plan based on these insights.
Only the votes surpassing five will be shown to the agents.
During inference on the validation set, agents are required to apply the insights learned during training. 
First, agents must select the insights they find helpful and then use them to guide their planning. 
Only insights with a vote count exceeding five are displayed to the agents for use during planning.

\subsection{Parametric Memory Updating}
\label{appendix-para-details}
\paragraph{OpenAI Models}
We use the official training script and default hyperparameters for OpenAI models. Specifically, for BlocksWorld, the hyperparameters are training steps set to 3, batch size set to 1, learning rate multiplier set to 2, and random seed set to 341541772.

For TravelPlanner, the hyperparameters are training steps set to 3, batch size set to 1, learning rate multiplier set to 2, and random seed set to 1294003109.

\paragraph{Open-source Models}
We fine-tune Llama3.1-8B and Qwen2-7B on 8$\times$A100 GPUs.
For BlocksWorld, the training step is 50, the batch size is 16, the learning rate is 1e-5, the learning rate schedule is cosine, and the warmup ratio is set to 0.1.

For TravelPlanner, due to the high computational cost associated with its longer context, we adopt LoRA as the training strategy. 
The training step is 200, the batch size is 2, and the learning rate is 1e-4. Other hyperparameters remain the same as in BlocksWorld.

\subsection{Model Access}
Our experiments utilize four closed-source LLMs accessed via API and five open-source LLMs. 
For the open-source models, we use instruction-tuned versions of each. 
Due to the high cost of deploying Llama3.1-405B, we perform inference on the Google Vertex AI platform and compute attribution scores locally. 
To ensure reproducibility, we have included the prompts used in our experiments in Appendix \ref{appendix:prompt-list}. 
For closed-source models, we use GPT-4o-2024-08-06, GPT-4o-mini-2024-07-18, o1-preview-2024-09-12, and o1-mini-2024-09-12 across all tests.

\subsection{Human-Written Insights}

We provide human-written insights for BlocksWorld and TravelPlanner here.

\lstset{
    backgroundcolor=\color[RGB]{245,245,245},
    breaklines=true,
    breakindent=0pt,
    basicstyle=\ttfamily\small,
    frame=trbl,
    frameround = tttt,
}\begin{lstlisting}
[BlocksWorld]
1. Only pick up or unstack one block at a time, ensuring your hand is empty before doing so.
2. A block can be picked up or unstacked only if it's clear and on the table.
3. A block is clear if it has no blocks on top and is not currently being held.
4. When unstacking, ensure the block you're removing is actually on top and clear.
5. After picking up or unstacking a block, you must hold it until it's placed down or stacked.
6. You can only place a block you're holding, and stacking can only occur if the target block is clear.
7. Once a block is placed down or stacked, your hand becomes empty, and the block below a newly stacked one is no longer clear.

[TravelPlanner]
1. Verify transportation and attraction availability before planning and provide alternatives if needed.
2. Ensure all plan details and activities are based on available data within the designated environment to avoid inaccuracies.
3. Include all essential details, such as accommodations and daily activities, ensuring they align logically with the planned city and timeline.
4. Maintain diversity by avoiding repetition of restaurant or attraction choices throughout the trip.
5. Ensure transportation methods are consistent and logical within the trip's context, avoiding conflicting options like self-driving and flights.
6. Follow any specified minimum night stay requirements when booking accommodations.
7. Plan activities, accommodations, and meals to align with the user's budget constraints.
8. Ensure accommodations comply with specific rules and preferences, including room type and restrictions on parties, smoking, pets, or visitors.
9. Adjust transportation options and other preferences according to the user's specified requirements, such as avoiding flights or self-driving.
10. Opt for budget-friendly accommodations, restaurants, and transportation methods.
\end{lstlisting}

\subsection{Attribution Score Calculation}
To obtain accurate attribution scores, we focus only on ``meaningful words'' in the analysis.
For instance, in BlocksWorld, we consider only actions such as ``Pick Up'', ``Put Down'', ``Stack'', and ``Unstack'', along with relevant objects like ``red block'', while discarding non-essential words like ``the''
A similar approach is applied to TravelPlanner, where only the values in the JSON format are considered. For example, in ``Accommodation: XXX'', only ``XXX'' is used for calculating attribution scores.

\subsection{Attribution Score Normalization}
Due to the varying planning steps and models, for example, the attribution scores in different models are in different scales, which cannot be compared directly.
To address this, we normalize the attribution scores by dividing each score by the maximum absolute value along the relevant dimension, which ensures that all scores are scaled consistently across different models. 
This normalization process allows for a fair comparison of the attribution scores by bringing them into a comparable range, typically between \num{-1} and \num{1}, without distorting the relative importance of features within the same model.

\newpage
\onecolumn

\section{Prompt List}
\label{appendix:prompt-list}
We provide the prompts utilized in this paper here.

\subsection{Behavioral Learning Prompt}

\lstset{
    backgroundcolor=\color[RGB]{245,245,245},
    breaklines=true,
    breakindent=0pt,
    basicstyle=\ttfamily\small,
    frame=trbl,
    frameround = tttt,
}\begin{lstlisting}
You are tasked with analyzing both successful and failed plans from previous attempts based on a specific query and background information. These failed plans are presented in chronological order, with the most recent plan including a detailed trajectory. As these plans fail to meet certain constraints, you are encouraged to refine the insights to improve them.

Use the following format to systematically analyze failed plans:
[State]: Describe the current situation, including factors like remaining budget, time constraints, and any other specified conditions in the query or provided information.
[Thought]: Explain the reasoning behind your decisions, considering the current state.
[Action]: Detail the specific parts of your plan in response to the [State] and [Thought].

For the successful plan, add a [Best Practice] section after the final analysis to summarize the key experiences and practices that led to success.
For the failed plan, add an [Error] section immediately after each defective [Action]. This section should identify and explain why the chosen actions or used insights were inappropriate, given the [State] and [Thought].

After evaluating the plans and previous insights, refine the current insight set based on findings from previous attempts and newly identified errors.

Your task involves adding, editing, supporting, and opposing insights from the existing set:
[Add]: Integrate new pairs that are missing in the current set. Add new ones only when absolutely necessary.
[Edit]: Revise pairs that are incomplete or partially incorrect. Editing an insight retains its number of votes.
[Support]: Endorsing specific pairs to emphasize their value. Increase the number of votes for the supported pair by 1 each time. Some previously used insights might have been edited (with the same index). If you support the new version, vote for it.
[Oppose]: Challenging insights that are incorrect, outdated, or only applicable under specific conditions. This will decrease the number of votes by 1 each time.

Opposing and editing are highly encouraged to resolve any conflicting insights. Avoid proposing insights with similar purposes.

Legal Action on Current Insight Set:
[Add/Edit/Support/Oppose] [Insight 1]: [Content].

Insight Set:
{insight_set}
-----
Task Instruction: {task}
-----

Successful Plan: 
{successful_plan}

Failed Plans:
{failed_plan}

Last Failed Plan Trajectory:
{trajectory}

Please use the following format for your response (do not output in the markdown style):
Successful Plan Analysis:
Failed Plan Analysis:
Action on Current Insight Set:
[Finished]
\end{lstlisting}

\subsection{Oracle Feedback Learning Prompt}

\lstset{
    backgroundcolor=\color[RGB]{245,245,245},
    breaklines=true,
    breakindent=0pt,
    basicstyle=\ttfamily\small,
    frame=trbl,
    frameround = tttt,
}\begin{lstlisting}
You are tasked with analyzing failed plans from previous attempts, along with their evaluation results, based on a specific query and background information. These failed plans are presented in chronological order, with the most recent plan including a detailed trajectory. As these plans fail to meet certain constraints, you are encouraged to refine the insights to improve them.

Use the following format to systematically analyze failed plans:
[State]: Describe the current situation, including factors like remaining budget, time constraints, and any other specified conditions in the query or provided information.
[Thought]: Explain the reasoning behind your decisions, considering the current state.
[Action]: Detail the specific parts of your plan in response to the [State] and [Thought].

For the failed plan, add an [Error] section immediately after each defective [Action]. This section should identify and explain why the chosen actions or used insights were inappropriate, given the [State] and [Thought].

After evaluating the plans and previous insights, refine the current insight set based on findings from previous attempts and newly identified errors.

Your task involves adding, editing, supporting, and opposing insights from the existing set:
[Add]: Integrate new pairs that are missing in the current set. Add new ones only when absolutely necessary.
[Edit]: Revise pairs that are incomplete or partially incorrect. Editing an insight retains its number of votes.
[Support]: Endorsing specific pairs to emphasize their value. Increase the number of votes for the supported pair by 1 each time. Some previously used insights might have been edited (with the same index). If you support the new version, vote for it.
[Oppose]: Challenging insights that are incorrect, outdated, or only applicable under specific conditions. This will decrease the number of votes by 1 each time.

Opposing and editing are highly encouraged to resolve any conflicting insights. Avoid proposing insights with similar purposes.
Note: Ensure that the insights are high-level and generalizable, rather than detailed and specific to particular queries. Make sure your contributions do not introduce unrelated insights or go beyond the scope of the provided information.

Legal Action on Current Insight Set:
[Add/Edit/Support/Oppose] [Insight 1]: [Content].

Insight Set:
{insight_set}
-----
Task Instruction: {task}
-----
Failed Plans:
{failed_plan}

Evaluation Results:
{eval_results}

Last Failed Plan Trajectory:
{trajectory}

Please use the following format for your response (do not output in the markdown style):
Failed Plan Analysis:
Action on Current Insight Set:
[Finished]
\end{lstlisting}

\newpage

\subsection{BlocksWorld Inference Prompt}

\lstset{
    backgroundcolor=\color[RGB]{245,245,245},
    breaklines=true,
    breakindent=0pt,
    basicstyle=\ttfamily\small,
    frame=trbl,
    frameround = tttt,
}\begin{lstlisting}
{query}

To help your plan, some insights from a set summarized by previous agents will be provided. Not all insights will be appropriate; you need to select the relevant ones to guide your plan. The values in brackets indicate the reliability of the insights, with higher values representing greater reliability.
Insight Set: {insight_set}

You should specify the insights you have chosen (beginning with [Chosen Insights]), followed by your final plan (beginning with [Plan]).
\end{lstlisting}

\subsection{TravelPlanner Inference Prompt}

\lstset{
    backgroundcolor=\color[RGB]{245,245,245},
    breaklines=true,
    breakindent=0pt,
    basicstyle=\ttfamily\small,
    frame=trbl,
    frameround = tttt,
}\begin{lstlisting}
You are a proficient planner. Based on the provided information and query, please give me a detailed plan, including specifics such as flight numbers (e.g., F0123456), restaurant names, and accommodation names. Note that all the information in your plan should be derived from the provided data. You must adhere to the format given in the example. Additionally, all details should align with commonsense. The symbol '-' indicates that information is unnecessary. For example, in the provided sample, you do not need to plan after returning to the departure city. When you travel to two cities in one day, you should note it in the 'Current City' section as in the example (i.e., from A to B).
Background Information:
{background information}
***** Example *****
Query: Please help me plan a trip from St. Petersburg to Rockford spanning 3 days from March 16th to March 18th, 2022. The travel should be planned for a single person with a budget of $1,700.

Plan:
[
{{
"days": 1,
"current_city": "from St. Petersburg to Rockford",
"transportation": "Flight Number: F3573659, from St. Petersburg to Rockford, Departure Time: 15:40, Arrival Time: 17:04",
"breakfast": "-",
"attraction": "-",
"lunch": "-",
"dinner": "Coco Bambu, Rockford",
"accommodation": "Pure luxury one bdrm + sofa bed on Central Park, Rockford"
}},
{{
"days": 2,
"current_city": "Rockford",
"transportation": "-",
"breakfast": "Flying Mango, Rockford",
"attraction": "Burpee Museum of Natural History, Rockford; Midway Village Museum, Rockford; Discovery Center Museum, Rockford",
"lunch": "Grappa - Shangri-La's - Eros Hotel, Rockford",
"dinner": "Dunkin' Donuts, Rockford",
"accommodation": "Pure luxury one bdrm + sofa bed on Central Park, Rockford"
}},
{{
"days": 3,
"current_city": "from Rockford to St. Petersburg",
"transportation": "Flight Number: F3573120, from Rockford to St. Petersburg, Departure Time: 19:00, Arrival Time: 22:43",
"breakfast": "Subway, Rockford",
"attraction": "Klehm Arboretum & Botanic Garden, Rockford; Sinnissippi Park, Rockford",
"lunch": "Cafe Coffee Day, Rockford",
"dinner": "Dial A Cake, Rockford",
"accommodation": "-"
}}
]
***** Example Ends *****
To help your plan, some insights from a set summarized by previous agents will be provided. Not all insights will be appropriate; you need to select the relevant ones to guide your plan. The values in brackets indicate the reliability of the insights, with higher values representing greater reliability.
Insight Set: {insight_set}
Given information: {text}
Query: {query}

You should specify the insights you have chosen (beginning with [Chosen Insights]), followed by your final plan (beginning with [Plan]).
\end{lstlisting}

\end{document}